\pdfoutput=1
\documentclass[11pt]{article}

\usepackage[preprint]{acl}

\usepackage{times}
\usepackage{latexsym}

\usepackage[T1]{fontenc}
\usepackage[utf8]{inputenc}

\usepackage{microtype}

\usepackage{inconsolata}

\usepackage{graphicx}

\usepackage{lipsum}
\usepackage{textcomp}  
\usepackage{scalerel} 
\usepackage{booktabs}
\usepackage{enumitem}
\usepackage{CJKutf8}
\usepackage{amssymb}
\usepackage{tabularx}
\usepackage{tablefootnote}
\usepackage{makecell}
\usepackage{caption}
\usepackage{subcaption}
\usepackage{dblfloatfix}

\usepackage[frozencache=true,cachedir=minted-cache]{minted}
\setminted{
frame=lines,
framesep=2mm,
baselinestretch=1.2,
fontsize=\footnotesize
}

\newcommand{\checkmrk}{\scalerel*{\includegraphics{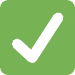}}{\textrm{\textbigcircle}}}
\newcommand{\cross}{\scalerel*{\includegraphics{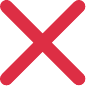}}{\textrm{\textbigcircle}}}
\newcommand{\paidmrk}{\scalerel*{\includegraphics{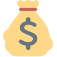}}{\textrm{\textbigcircle}}}

\title{ReDel: A Toolkit for LLM-Powered Recursive Multi-Agent Systems}

\author{
    Andrew Zhu, \quad Liam Dugan, \quad Chris Callison-Burch \\
    University of Pennsylvania \\
    {\tt \{andrz,ldugan,ccb\}@seas.upenn.edu}
}

\usepackage{xcolor}
\usepackage{xifthen}
\newcommand\todo[1]{
    \textcolor{red}{
        \ifthenelse{\isempty{#1}}
            {TODO}
            {TODO: #1}
    }
}

\begin{document}
\maketitle

\begin{abstract}
Recently, there has been increasing interest in using Large Language Models (LLMs) to construct complex multi-agent systems to perform tasks such as compiling literature reviews, drafting consumer reports, and planning vacations. Many tools and libraries exist for helping create such systems, however none support \textit{recursive} multi-agent systems---where the models themselves flexibly decide when to delegate tasks and how to organize their delegation structure. In this work, we introduce ReDel: a toolkit for recursive multi-agent systems that supports custom tool-use, delegation schemes, event-based logging, and interactive replay in an easy-to-use web interface. We show that, using ReDel, we are able to easily identify potential areas of improvements through the visualization and debugging tools. Our code, documentation, and PyPI package are open-source\footnote{ReDel's source code is available at \url{https://github.com/zhudotexe/redel}.} and free to use under the MIT license.
\end{abstract}

\section{Introduction}
A multi-agent system uses multiple large language models (LLMs) together to accomplish complex tasks or answer complex questions beyond the capabilities of a single LLM. Often, in such scenarios, each LLM is provided with tools \cite{parisi2022talmtoolaugmentedlanguage, schick2023toolformer} that it can use to give it additional capabilities, like searching the internet for real-time data or interacting with a web browser. 
In most cases, these systems are defined manually, with a human responsible for defining a static problem-decomposition graph and defining an agent to handle each subproblem in the graph \cite[\textit{inter alia}]{hong2023metagptmetaprogrammingmultiagent, wu2023autogenenablingnextgenllm, zhang2024proagentbuildingproactivecooperative, qiao2024autoactautomaticagentlearning}.

\begin{figure}[t]
    \centering
    \includegraphics[width=\columnwidth]{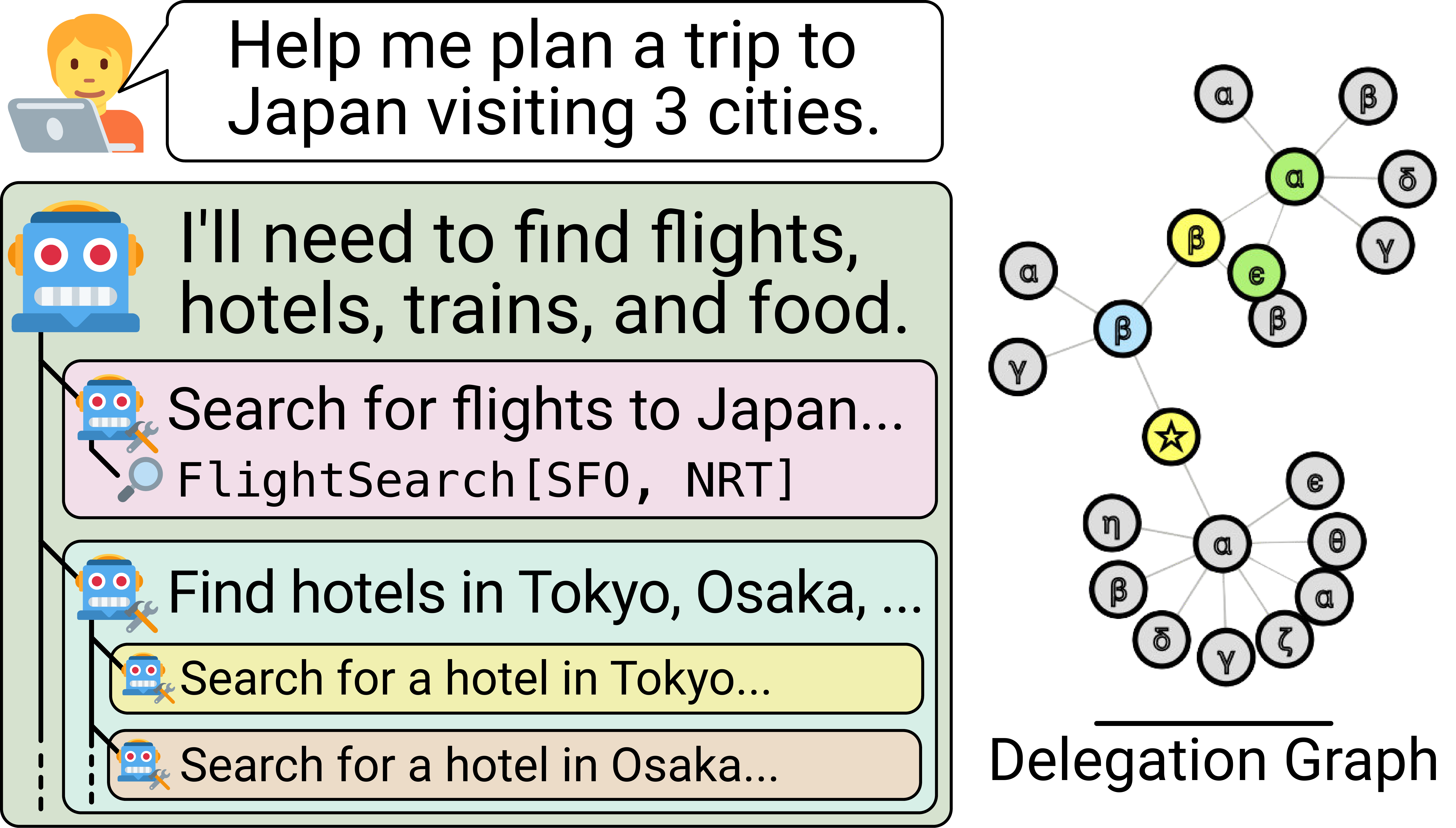}
    \caption{ReDel allows developers to create systems of recursive agents, inspect each agent's state, and visualize a system's delegation graph (right). Recursive agents can be used to solve complex tasks, such as planning a trip to Japan (left).}
    \label{fig:fig1}
\end{figure}

In a \textit{recursive} multi-agent system, rather than a human defining the layout of multiple agents, a single root agent is given a tool to spawn additional agents. When faced with a complex task, the root agent can decompose the task into smaller subtasks, then delegate those tasks to newly-created sub-agents. Each sub-agent can then either complete the task if it is small enough, or recursively decompose and delegate the task further\footnote{This is where the toolkit’s name, ReDel, comes from: it’s short for \textbf{Re}cursive \textbf{Del}egation.} \cite{khot2023decomposedpromptingmodularapproach, lee-kim-2023-recursion, prasad-etal-2024-adapt} (Figure \ref{fig:fig1}).

In the current landscape of multi-agent systems, the majority of tooling focuses on human-defined static systems, and poorly handles dynamic systems where agents are added to a computation graph at runtime. Furthermore, much of this tooling is unsuitable for academic purposes \cite{zhu-etal-2023-kani} or hidden behind paywalls and proprietary licenses. 

In this paper, we present ReDel, a fully-featured open-source toolkit for recursive multi-agent systems.
ReDel makes it easy to experiment by providing a \textbf{modular interface} for creating tools, different delegation methods, and logs for later analysis. This granular logging and a central \textbf{event-driven system} makes it easy to listen for signals from anywhere in a system, and every event is automatically logged for post-hoc data analysis. ReDel also features a \textbf{web interface} that allows users to interact with a configured system directly and view replays of saved runs, making it easy for researchers and developers to build, iterate on, and analyze recursive multi-agent systems. In Section \ref{sec:experiments} we use ReDel to run recursive multi-agent systems on three diverse agentic benchmarks, and in Section \ref{sec:error_analysis} we demonstrate how the toolkit can be used to explore complex behaviours of these systems.

\section{Related Work}

\paragraph{Recursive Multi-Agent Systems.}
Recent work on recursive multi-agent systems has been done by \citet{lee-kim-2023-recursion}, \citet{khot2023decomposedpromptingmodularapproach}, \citet{qi-etal-2023-art}, and \citet{prasad-etal-2024-adapt}. These works introduce the method of fine-tuning or few-shot prompting LLMs to decompose complex tasks and using sub-agents to solve each part (often called recursive or hierarchical decomposition). 
ReDel builds upon the methods introduced in these works by taking advantage of modern models' native tool use capability \cite{schick2023toolformer} to decompose and delegate tasks zero-shot (i.e., without human-written examples in prompt) instead of using few-shot prompting or fine-tuning. As a framework, we provide an extensible interface to apply these approaches to additional tasks and domains.

Other multi-agent system methods such as agent evolution \cite{qian2024investigateconsolidateexploitgeneralstrategyintertask, yuan2024evoagentautomaticmultiagentgeneration, zhou2024symboliclearningenablesselfevolving} perturb human-written prompts and tools to create  new variations of sub-agents on the fly. In this paper, we choose to explore delegation using zero-shot prompting and function calling without on-the-fly adaptation, but our framework is flexible enough to implement these alternate methods of agent delegation as well.

\paragraph{Multi-Agent System Frameworks.}
Although there are other LLM-powered multi-agent system frameworks, each have various weaknesses that make them poorly suited for recursive systems and/or academic purposes. In Table \ref{tab:feature_comparison}, we compare LangGraph \cite{campos2023langgraph}, LlamaIndex \cite{Liu_LlamaIndex_2022}, MetaGPT \cite{hong2023metagptmetaprogrammingmultiagent}, AutoGPT \cite{Significant_Gravitas_AutoGPT}, and XAgent \cite{xagent2023} to ReDel, our system.
Most are built around static multi-agent systems, with only AutoGPT and XAgent supporting a single level of delegation.
Only LangGraph and LlamaIndex allow agents to run in parallel asynchronously, whereas MetaGPT, AutoGPT, and XAgent run one agent at a time in a synchronous fashion.
To log events deep within the system, only LlamaIndex provides a rigorous instrumentation suite to developers that allows them to emit events at any point while a system is running.
Most do not allow developers to replay a system run from a log, with only LangGraph allowing replays by taking snapshots of each state of the system. 
Most do not provide a visualization interface, with only AutoGPT and XAgent providing a simple chat-based UI. Unless one subscribes to a paid service, LangGraph's replays cannot be viewed visually, and are instead presented as the raw data of each state.
Finally, only AutoGPT, MetaGPT, and XAgent are fully open-source, with LangGraph and LlamaIndex utilizing proprietary code to offer more ``premium'' features beyond what their open-source libraries offer.

In comparison, ReDel allows developers to customize their agents' delegation strategies and build multi-level dynamic systems while providing all of these features out of the box and remaining fully free and open source. It is the only such toolkit to provide first-class support for recursive multi-agent systems with best-in-class support for system visualization and modern LLMs with tool usage.

\begin{table}
    \small
    \centering
    \begin{tabular}{l|cccccc}
    \toprule
    & \rotatebox[origin=c]{-90}{ReDel} & \rotatebox[origin=c]{-90}{LangGraph} & \rotatebox[origin=c]{-90}{LlamaIndex} & \rotatebox[origin=c]{-90}{MetaGPT} & \rotatebox[origin=c]{-90}{AutoGPT} & \rotatebox[origin=c]{-90}{XAgent} \\
    \midrule
    Dynamic Systems & \checkmrk & \cross & \cross & \cross & \checkmrk & \checkmrk \\
    Parallel Agents & \checkmrk & \checkmrk & \checkmrk & \cross & \cross & \cross \\
    Event-Driven & \checkmrk & \cross & \checkmrk & \cross & \cross & \cross \\
    Run Replay & \checkmrk & \checkmrk & \cross & \cross & \cross & \cross \\
    Web Interface & \checkmrk & \paidmrk & \cross & \cross & \checkmrk & \checkmrk \\
    Fully Open Source & \checkmrk & \cross & \cross & \checkmrk & \checkmrk & \checkmrk \\
    \bottomrule
    \end{tabular}
    \caption{A feature comparison between ReDel and competing toolkits. ReDel is the only fully open-source toolkit that supports dynamic multi-agent systems with a rich event-driven base and web interface.}
    \label{tab:feature_comparison}
\end{table}

\section{System Design}
ReDel consists of two main parts: a Python package to define recursive delegation systems, log events, and run experiments, and a web interface to quickly and interactively iterate on defined systems or analyze experiment logs. In the following sections, we discuss these components in more detail.

\subsection{Tool Usage}
In ReDel, a ``tool'' is a group of functions, written in Python, that is exposed to an agent. The agent may generate requests to call appropriate functions from this tool, which interact with the environment (e.g. searching the Internet).

\begin{figure}
\begin{minted}{python}
class MyHTTPTool(ToolBase):
  @ai_function()
  def get(self, url: str):
    """Get the contents of a webpage,
    and return the raw HTML."""
    resp = requests.get(url)
    return resp.text
\end{minted}
\caption{An example of a simple ReDel tool that exposes an HTTP GET function to any agent equipped with the tool.}
\label{fig:tool_example}
\end{figure}

Developers can define tools in any Python file, and a tool's methods can be implemented by any Python code. ReDel is implemented in pure Python, and method bodies will not be sent to an agent's underlying language model, so there is no limit to a tool's implementation complexity or length. Similarly, a tool can use functionality defined in any other external library, allowing developers to utilize existing application code. An example of a basic tool that provides a function for making HTTP requests is in Figure \ref{fig:tool_example}.

ReDel comes bundled with a web browsing tool and email tool as examples, and we encourage developers to implement domain-specific tools for their own purposes.

\begin{figure}
\begin{minted}{python}
prompt_toks = Counter()
out_toks = Counter()

for event in read_jsonl("/path/to/events.jsonl"):
  if event["type"] == "tokens_used":
    eid = event["id"]
    prompt_toks[eid] += event["prompt_tokens"]
    out_toks[eid] += event["completion_tokens"]
\end{minted}
\caption{Every event in a ReDel system, builtin or custom, is logged to a JSONL file. Developers can use data analysis tools of their choice to analyze event logs post-hoc. This example demonstrates token counting.}
\label{fig:event_usage}
\end{figure}

\begin{figure}[h]
\begin{minted}{python}
# define a custom event
class CustomToolEvent(BaseEvent):
  type: Literal["custom_event"] = "custom_event"
  id: str  # the ID of the dispatching agent 
  foo: str  # some other data

# define a tool that dispatches the event
class MyTool(ToolBase):
  @ai_function()
  def my_cool_function(self):
    self.app.dispatch(
      CustomToolEvent(id=self.kani.id, foo="bar")
    )
    # other behaviour here ...
\end{minted}
\caption{Using ReDel to define a custom event and dispatch it from a tool. Custom events can be used to add observability deep within a system and can be queried post-hoc for rich data analysis.}
\label{fig:event_example}
\end{figure}

\subsection{Delegation Schemes}
A delegation scheme is the strategy used by an agent to send tasks to sub-agents. In ReDel, delegation schemes are implemented as a special type of tool that an LLM agent (the ``parent'') can call with task instructions as an argument. These instructions are sent to a new sub-agent (the ``child''), which can either complete them if they are simple enough, or break them up into smaller parts and recursively delegate again.

\begin{figure*}[!b]
    \centering
    \begin{subfigure}[t]{\columnwidth}
        \centering
        \includegraphics[width=\columnwidth]{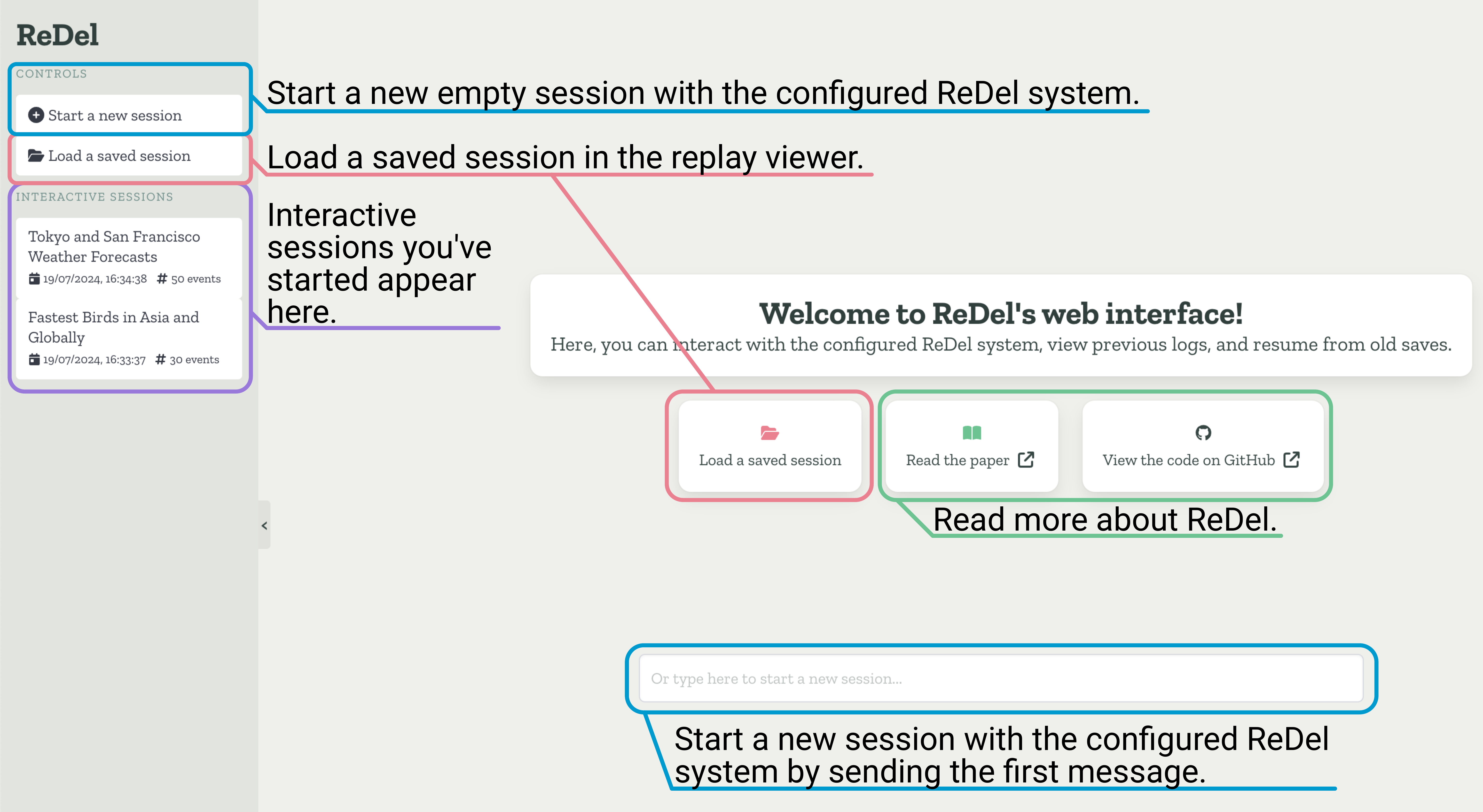}
        \caption{The \textbf{home page} of the ReDel web interface.}
        \label{fig:viz_home}
    \end{subfigure}
    ~
    \begin{subfigure}[t]{\columnwidth}
        \centering
        \includegraphics[width=\columnwidth]{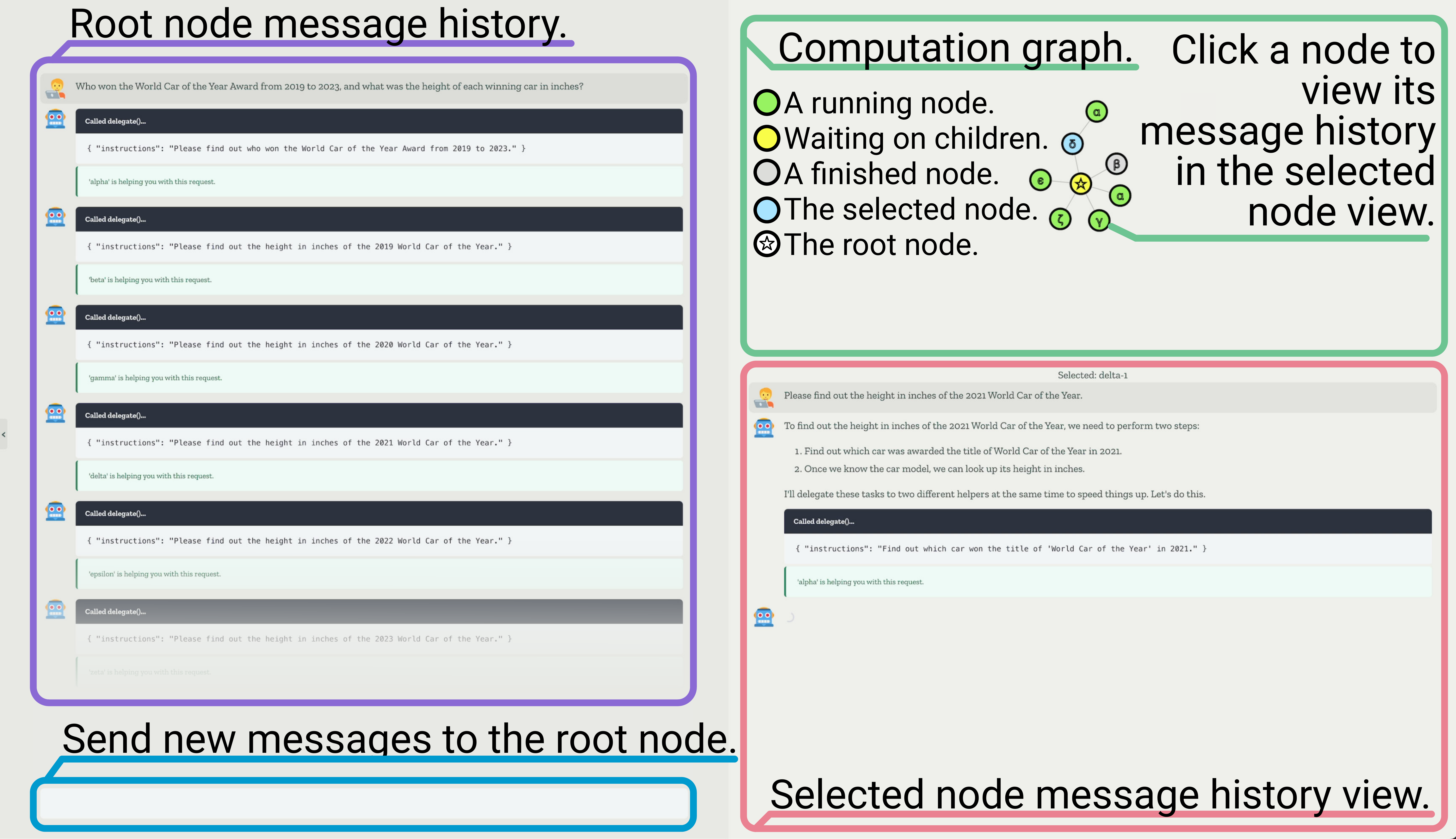}
        \caption{ReDel's \textbf{interactive view} allows users to quickly iterate on prompts and tool design, and test end-to-end performance.}
        \label{fig:viz_interactive}
    \end{subfigure}
    
    \begin{subfigure}[t]{\columnwidth}
        \centering
        \includegraphics[width=\columnwidth]{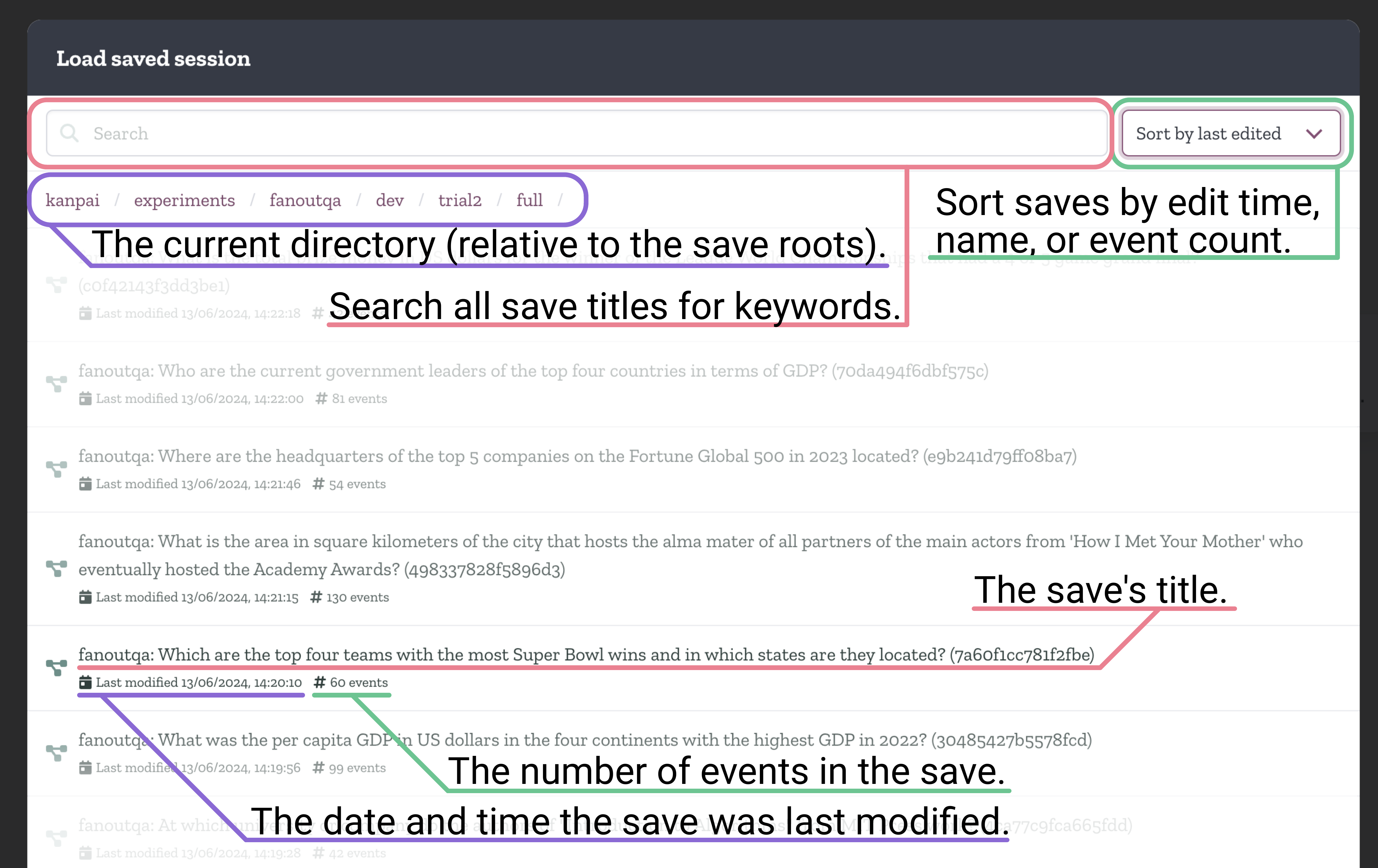}
        \caption{The \textbf{save browser} displays logs found in configured directories on the filesystem. It allows developers to search for and review previous runs of ReDel systems.}
        \label{fig:viz_save_browser}
    \end{subfigure}
    ~
    \begin{subfigure}[t]{\columnwidth}
        \centering
        \includegraphics[width=\columnwidth]{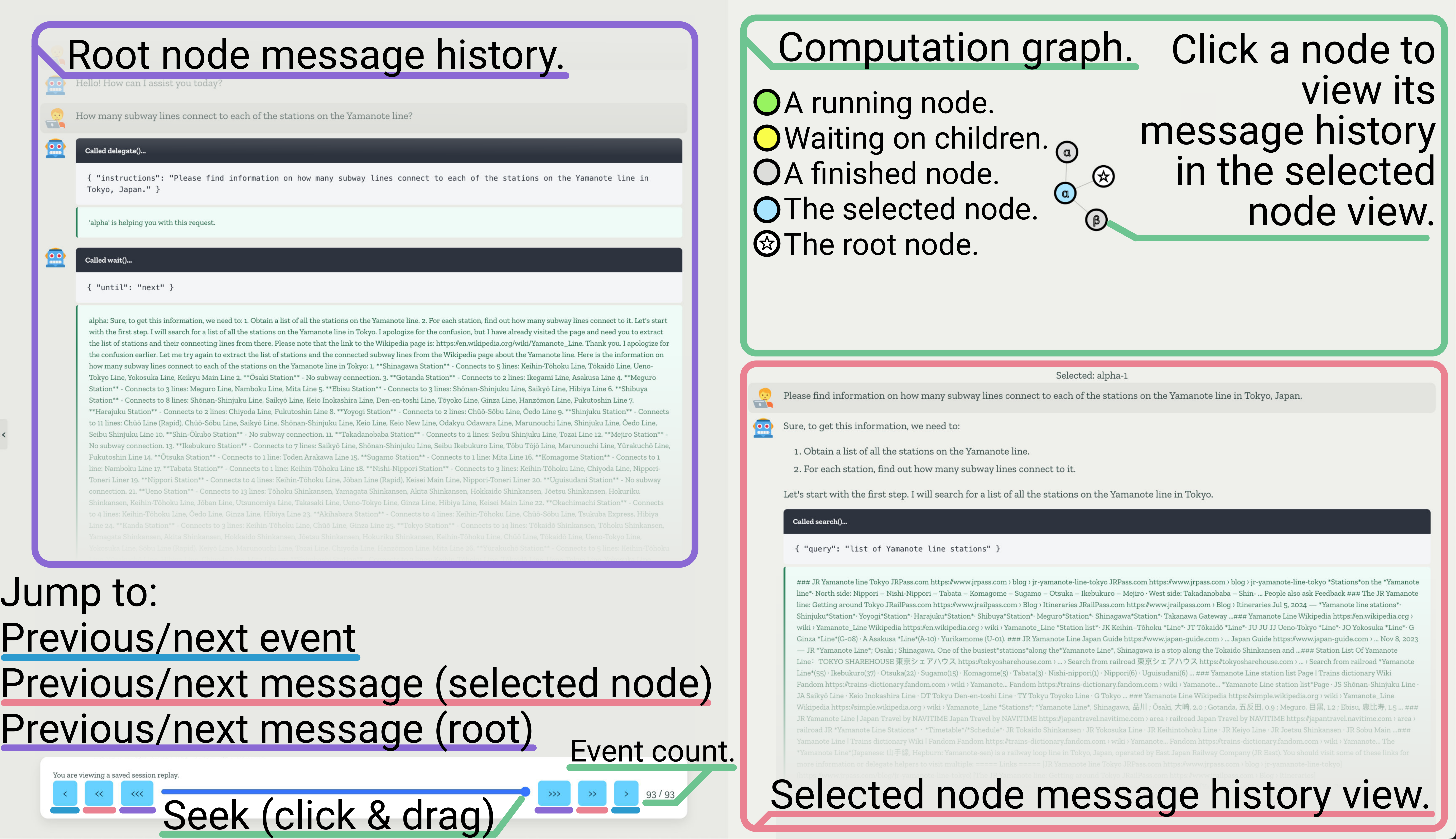}
        \caption{ReDel's \textbf{replay view} allows developers to replay saved runs of ReDel systems, giving events temporal context when analyzing or debugging a system's performance.}
        \label{fig:viz_replay}
    \end{subfigure}

    \caption{The four views of the ReDel web interface: Home (a), Interactive (b), Save Browser (c), and Replay (d).}
    \label{fig:viz}
\end{figure*}

Taking inspiration from common process management paradigms found in operating systems, ReDel comes with two delegation schemes: 
\begin{itemize}[noitemsep]
    \item \textbf{\texttt{DelegateOne}}: \textit{Synchronously} block the parent agent's execution until the child agent returns its result (in the form of its chat output).
    \item \textbf{\texttt{DelegateWait}}\footnote{Named so in that it provides two functions to agents: \texttt{delegate()}, which sends the instructions to the child agent and spawns it, and \texttt{wait()}, which retrieves its result, waiting for it to finish if necessary.}: Do not block parent agent's execution. Instead, provide a separate function to \textit{asynchronously} retrieve the result (chat output) of a particular child.
\end{itemize}
The \texttt{DelegateOne} scheme is well-suited for LLMs with parallel function calling as it allows ReDel to let a group of spawned child agents run in parallel, and return their results once they all complete. 

In contrast, the \texttt{DelegateWait} scheme is well-suited for LLMs without parallel function calling, as it lets these models spawn multiple agents before deciding to wait on any one agent's result (i.e., retrieve its conversational output). The drawback is that this runs the risk of creating zombie agents if the parent agent never retrieves the results of a particular child agent.\footnote{From our testing, this is a fairly rare occurrence.} As far as we are aware, ReDel is the first system to implement this type of deferred delegation scheme.

Developers can also implement their own delegation schemes modularly in a fashion similar to defining tools which can enable more complex behaviour. For example, a developer might implement a delegation scheme that allows a parent agent to ask follow-up questions to existing children to enable multi-turn delegation. Developers can also use the delegation scheme to control how the child passes information back to its parent -- for example, having each child call a \texttt{set\_result()} function to explicitly record its answer to a subtask instead of implicitly sending its chat output to the parent. We include examples of how to define a delegation scheme in Appendix \ref{sec:additional_code_examples} and in our GitHub repository.

\subsection{Events \& Logging}
ReDel operates as an event-driven framework, with comprehensive built-in events and the ability to define custom events. An event can be defined as anything from the creation of a sub-agent to the usage of a particular tool. Whenever ReDel catches an event, it logs the event to a JSONL file. This file essentially acts as an execution trace for a system run and users can use standard data analysis tools to inspect this trace and debug their runs. Figure \ref{fig:event_usage} shows how a basic Python script can be used to count a system's token usage post-hoc. 

Furthermore, using just the built-in events, ReDel is able to interactively play back any response through our web interface for extra visual debugging aid (see Section \ref{sec:web-interface}). In Section \ref{sec:experiments} we show a case study of how this can be used to debug complex query failures. We provide the set of built-in default events in Appendix \ref{sec:stock_events} and an example of defining a custom event in Figure \ref{fig:event_example}.

\subsection{Web Interface}
\label{sec:web-interface}
The web interface consists of four main views:

\paragraph{Home Page.} 
The home page (Figure \ref{fig:viz_home}) is the default view when starting the interface for the first time. Users can transition to the interactive view by sending a message in the chat bar, or use the provided buttons to load a saved replay or read more about ReDel. The sidebar lets users switch between interactive sessions they have started, start new sessions, or load saved replays.

\paragraph{Interactive View.}
In the interactive view (Figure \ref{fig:viz_interactive}), users can send messages to the root node to interact with the system. While the system is running, the top right panel contains the delegation graph: a visual representation of each agent in the system, their parent and children, and what their current status is: running (green), waiting (yellow), or done (grey). Users can further inspect each node in the delegation graph by clicking it, which displays its full message history in the bottom right panel. ReDel supports streaming, and LLM generations appear in real-time for every agent.

\begin{table*}[t]
\centering
\small
\begin{tabular}{l|cc|ccc|ccc}
\toprule
& \multicolumn{2}{c|}{\textbf{FanOutQA}} & \multicolumn{3}{c|}{\textbf{TravelPlanner}} & \multicolumn{3}{c}{\textbf{WebArena}} \\
\textbf{System}                 & \textbf{Loose} & \textbf{Model Judge}  & \textbf{CS-Micro} & \textbf{H-Micro} & \textbf{Final}    & \textbf{SR} & \textbf{SR (AC)} & \textbf{SR (UA)} \\
\midrule
ReDel (GPT-4o)                  & \textbf{0.687} & \textbf{0.494} & \textbf{67.49} & 9.52 & \textbf{2.78} & 0.203 & \textbf{0.179} & 0.643 \\
ReDel (GPT-3.5-turbo)  & 0.300          & 0.087          & 54.58 & 0 & 0 & 0.092 & 0.066 & 0.571 \\
\midrule
Baseline (GPT-4o)               & 0.650          & 0.394     & 50.83 & \textbf{18.81} & 0 & 0.162 & 0.128 & \textbf{0.786} \\
Baseline (GPT-3.5-turbo)        & 0.275          & 0.077          & 48.75 & 0.24 & 0 & 0.085 & 0.058 & 0.571 \\
\midrule
Published SotA & 0.580 & 0.365 & 61.1 & 15.2 & 1.11 & \textbf{0.358} & --- & --- \\
\bottomrule
\end{tabular}
\caption{Systems' performance on FanOutQA, TravelPlanner, and WebArena. The SotA models are GPT-4o on FanOutQA, GPT-4-turbo/Gemini Pro on TravelPlanner, and SteP on WebArena. We see that ReDel outperforms the corresponding single-agent baselines across all benchmarks and improves over published SotA in two of three.}
\label{tab:results}
\end{table*}

\paragraph{Save Browser.}
The save browser (Figure \ref{fig:viz_save_browser}) allows users to select replays to view from the list of previous sessions. This allows researchers to run experiments in batches while saving their logs, and use the interface to review the system's behaviour at a later date. The save list contains all the saves that the ReDel server found in the provided save directories, their titles, number of events, and when they were last edited. Users can search for keywords in a save's title and can also sort saves by name, edit time, or number of events -- the latter allowing users to quickly find outliers at a glance.

\paragraph{Replay View.}
With just the built-in default events (see Appendix \ref{sec:stock_events}) ReDel saves enough information about a session to fully recreate it in a replay setting. Thus, the replay view (Figure \ref{fig:viz_replay}) allows users to step through every event (both built-in and custom) dispatched by the system during a particular session and visualize each event's impact on the system.

The layout of the replay view is virtually identical to the interactive view except with the message bar replaced by replay controls. Users can use these controls to jump between messages in the root node, selected node in the delegation graph, or seek events using the slider. The message history and delegation graph update in real time as users seek through the replay.

\section{Evaluation \& Case Study}
\label{sec:experiments}
To evaluate ReDel, we compare its performance to a baseline single-agent system and to the published state-of-the-art system on three different benchmarks. We include the logs and source code for all experiments in our code release.

\subsection{Experimental Setup}

\paragraph{Benchmarks.}
To properly evaluate ReDel we had to choose only datasets that contained sufficiently complex tasks. For our benchmarks we therefore chose the following:
\begin{enumerate}[noitemsep]
    \item \textbf{FanOutQA}: \cite{zhu2024fanoutqamultihopmultidocumentquestion} Agents must compile data from many Wikipedia articles to answer complex information-seeking queries.
    \item \textbf{TravelPlanner}: \cite{xie2024travelplanner} Agents must create travel plans using tools to search flights, restaurant, and attraction databases.
    \item \textbf{WebArena}: \cite{zhou2024webarena} Agents must do complex web tasks such as adding products to a shopping cart or commenting on GitLab.
\end{enumerate}
Due to cost constraints we limited our evaluation to roughly 100-300 examples from each benchmark (see Appendix \ref{sec:experiment_configs}).

\paragraph{Models.}
For our main two ReDel systems we used GPT-4o \cite{gpt-4o} and GPT-3.5-turbo \cite{gpt-35-turbo} as the underlying models. In all setups, root nodes are not given tool usage capabilities and use the DelegateOne delegation scheme.

For the two baseline systems, we used the GPT-4o and GPT-3.5-turbo models as-is. All models were given equal access to all tools and no few-shot prompting or fine-tuning was performed. 

\subsection{Results}

In Table \ref{tab:results} we report the results of our evaluation. We see that, across all benchmarks, our recursive delegation system significantly outperforms its corresponding single-agent baseline. We even present an improvement over the previous state of the art systems in both FanOutQA and TravelPlanner.

Furthermore, we see that the gap between ReDel and the baseline system gets larger as the capabilities of the underlying model improves. We believe that this bodes well for the application of such techniques to future, more powerful models.

In the few cases where ReDel fails, namely H-Micro on TravelPlanner and SR on WebArena, these are attributable to metric failures and unequal comparisons. In the TravelPlanner case, on further inspection, we find that recursive systems tend to make more commonsense inputs for meals (e.g. ``on the flight'' or ``packed lunch'') -- which causes the TravelPlanner evaluation script to give a score of 0 on the Hard Constraint metric. As for the WebArena result, the published SotA SteP model uses few-shot, chain-of-thought prompting, whereas our systems all use zero-shot prompting.

\section{Using ReDel for Error Analysis}
\label{sec:error_analysis}
For our error analysis, we took the saved log files for each benchmark and manually investigated the logs of both the successful runs as well as the failed runs through the replay view of the ReDel web interface. Through this investigation we  observed two common failure cases in recursive multi-agent systems. These cases are as follows: 
\begin{itemize}[noitemsep]
\item \textbf{Overcommitment}: The agent attempts to complete an overly-complex task itself.
\item \textbf{Undercommitment}: The agent performs no work and re-delegates the task it was given.
\end{itemize}
We find that overcommitment commonly occurs when an agent performs multiple tool calls and fills its context window with retrieved information. 
In the ReDel web interface, this manifests as an abnormally small delegation graph, often consisting of only two nodes: the root node, and a single child which the root delegates to and which subsequently overcommits. In practice, this often, but not always, results in the overcommitting model ``forgetting'' the task it was meant to accomplish due to the original task being truncated its limited context window. An overcommitting model might fail a task because it outputs a summary of whatever remains in its context window instead of the answer to the original task, whereas a task failure due to causes other than overcommittment might look like a hallucinated result or a simple apology for being unable to complete the task.

\begin{figure}[t]
    \centering
    \includegraphics[width=\columnwidth]{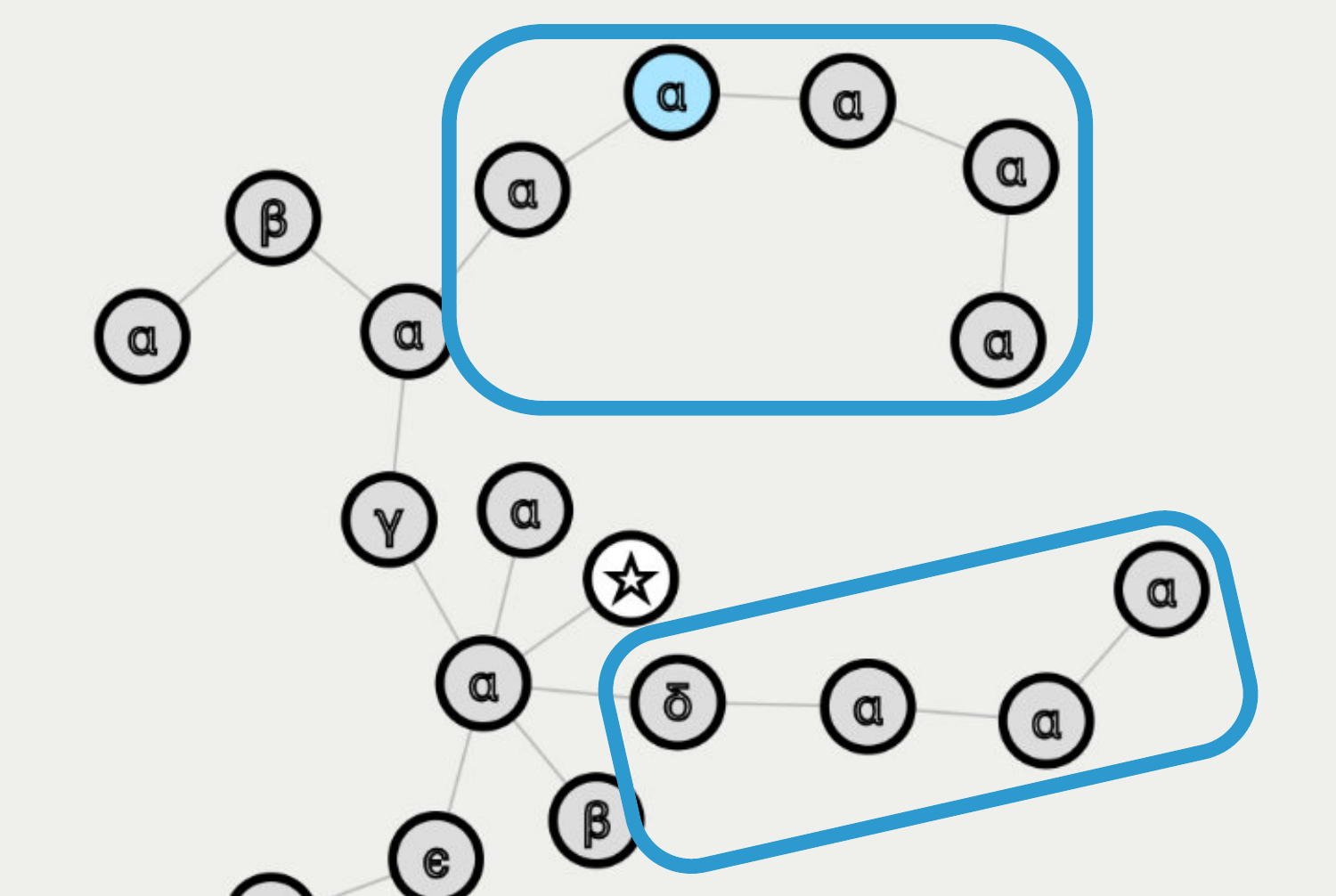}
    \caption{Recursive systems exhibiting undercommitment produce long chains of agents (blue boxes), as seen in the ReDel delegation graph.}
    \label{fig:undercommitment}
\end{figure}

In contrast, we find that undercommitment commonly happens when the model incorrectly decides that it does not have the necessary tools to solve the problem and instead assumes that its future child will possess the required tools to solve the problem. In all three benchmarks, this led to failure as agents entered an infinite loop of delegation until they reached a configured depth limit or timed out. In the web interface, this manifests as a line of nodes in the delegation graph (Figure \ref{fig:undercommitment}).

In Table \ref{tab:commitment} we tabulate the over- and undercommitment rates of ReDel with both GPT-4o and GPT-3.5-turbo for each benchmark. We did this heuristically by counting any delegation graph with two or fewer agents as overcommitted and any delegation graph with a chain of three or more agents with exactly zero or one children as undercommitted. We see that as models get stronger they have a stronger propensity to delegate. However, that propensity to delegate may lead to undercommitment.

\begin{table}
\centering
\small
\begin{tabular}{l|cc|cc|cc}
\toprule
& \multicolumn{2}{c|}{\textbf{FOQA}} & \multicolumn{2}{c|}{\textbf{TP}} & \multicolumn{2}{c}{\textbf{WA}} \\
\textbf{System} & \textbf{OC} & \textbf{UC} & \textbf{OC} & \textbf{UC} & \textbf{OC} & \textbf{UC} \\
\midrule
RD (4o) & 22.7 & 11.3 & 41.1 & 0.5 & 31.3 & 44.8 \\
RD (3.5-t) & 40.8 & 1.1 & 96.7 & 0 & 54.6 & 17.7 \\
\bottomrule
\end{tabular}
\caption{The overcommitment (OC) and undercommitment (UC) rates, in percent, of the two recursive multi-agent systems we tested, by benchmark.}
\label{tab:commitment}
\end{table}

Given the prevalence of these two issues, we hypothesize that recursive multi-agent systems may still see further improvements to performance from interventions that target these behaviors. For example, one could fine-tune or prompt agents with domain-specific instructions that detail when the models should delegate and when they should perform tasks on their own.

While implementing such improvements is beyond the scope of this paper, we believe that this case study helps to demonstrate the strengths of the ReDel system. Using the delegation graph view, it is easy to identify and characterize errors in recursive multi-agent systems and we hope that through ReDel more research can be done to further refine such systems for maximum utility.

\section{Conclusion}
We present ReDel, a novel toolkit for working with recursive multi-agent systems. ReDel allows academic developers to quickly build, iterate on, and run experiments involving dynamic multi-agent systems. It offers a modular interface to create tools for agents to use, an event framework to instrument experiments for later analysis, and a free and open-source web interface to interact with and explore developer-defined systems. We use ReDel to demonstrate recursive multi-agent systems' performance on three diverse benchmarks, and we include the full logs of these runs in our demo release for reproducibility and further exploration\footnote{\url{https://datasets.mechanus.zhu.codes/redel-dist.zip}}. 
ReDel opens the door for a new paradigm of recursive multi-agent systems, and we are excited to see how developers can utilize our system in the future.

\section*{Acknowledgements}

This research is supported in part by the Office of the Director of National Intelligence (ODNI), Intelligence Advanced
Research Projects Activity (IARPA), via the HIATUS Program contract \#2022-22072200005.
This material is based upon work supported by the National Science Foundation Graduate Research Fellowship, under Grant No. DGE-2236662.
The views and conclusions contained herein are those of the authors and should not be interpreted as necessarily
representing the official policies or views, either expressed or implied, of ODNI, IARPA, the NSF, or the U.S. Government. The
U.S. Government is authorized to reproduce and distribute reprints for governmental purposes notwithstanding any
copyright annotation therein.

\bibliography{custom}

\appendix
\onecolumn

\section{Custom Delegation Scheme}
\label{sec:additional_code_examples}

The following annotated code snippet shows how to use the ReDel Python package to define a delegation scheme -- the delegation scheme here is a reproduction of the bundled \texttt{DelegateOne} scheme.

\begin{figure}[h]
\begin{minted}{python}
class DelegateOne(DelegationBase):
    @ai_function()
    async def delegate(instructions: str):
        """(Insert your prompt for the model here.)"""

        # request a new agent instance from the system
        subagent = await self.create_delegate_kani(instructions)

        # set the state of the delegator agent to be waiting on the delegate
        with self.kani.run_state(RunState.WAITING):
            # buffer the delegate's response as a list of strings, filtering for ASSISTANT messages
            # use full_round_stream so that the app automatically dispatches streaming events
            result = []
            async for stream in subagent.full_round_stream(instructions):
                msg = await stream.message()
                if msg.role == ChatRole.ASSISTANT and msg.content:
                    result.append(msg.content)

            # clean up any of the delegate's ephemeral state and return result to caller
            await subagent.cleanup()
            return "\n".join(result)
\end{minted}
\caption{Using ReDel to define a custom delegation scheme. Delegation tools are responsible for the lifecycle of any agent they create.}
\label{fig:delegation_example}
\end{figure}

\section{Application Events}
\label{sec:stock_events}

The following table lists the built-in default events that will be emitted on every run of a ReDel system. Each event has a \texttt{type} key which is used to determine what kind of event it is, and a \texttt{timestamp} key.

\begin{table}[h]
\small
\centering
\renewcommand{\arraystretch}{1.2}
\begin{tabularx}{\textwidth}{ccXX}
\toprule
\textbf{Event Name} & \textbf{Key} & \textbf{Description} \\
\midrule
Agent Spawned & kani\_spawn & A new agent was spawned. The data attached to the event contains the full state of the agent at the time it was spawned, which includes its ID, relations to other agents, a description of the LLM powering it, the tools it has access to, and any system prompts. \\
Agent State Change & kani\_state\_change & The running state of an agent changed (e.g. from RUNNING to WAITING). Contains the ID of the agent and its new state. \\
Tokens Used & tokens\_used & An agent made a call to the language model powering it. Contains the ID of the agent, the number of tokens in the prompt it sent, and the number of tokens in the completion the LLM returned. \\
Agent Message & kani\_message & An agent added a new message to its chat history. Contains the ID of the agent and the message's role (e.g. USER or ASSISTANT) and content. \\
Root Message & root\_message & Similar to Agent Message, but only fires for messages in the root node. This is fired in addition to an Agent Message event. \\
Round Complete & round\_complete & Fired when the root node completes a full chat round (i.e. there are no running children and it has generated a response to a user query). \\
\bottomrule
\end{tabularx}
\caption{A list of events built-in to the ReDel toolkit.}
\label{tab:stock_events}
\end{table}

\section{Benchmark Comparison}
\label{sec:experiment_configs}
Here, we tabulate each of the benchmarks tested in our experiments.

\begin{table}[h]
\small
\centering
\renewcommand{\arraystretch}{1.2}
\begin{tabularx}{\textwidth}{p{0.15\linewidth}|ccp{0.17\linewidth}X}
\toprule
\textbf{Benchmark} & \textbf{Split} & \textbf{\#} & \textbf{Example} & \textbf{Metrics} \\
\midrule
\textbf{FanOutQA}\newline \cite{zhu2024fanoutqamultihopmultidocumentquestion}      & dev              & 310 & What is the total number of employees in the five largest banks in the world? &\textbf{Loose}: The average proportion of reference strings found in the generated answer.\newline \textbf{Model Judge}: Whether the reference answer and generated answer are equivalent, judged by GPT-4 (\texttt{gpt-4-0613}).\\
\midrule
\textbf{TravelPlanner}\newline \cite{xie2024travelplanner} & val& 180 & Please help me plan a trip from St. Petersburg to Rockford spanning 3 days from March 16th to March 18th, 2022. The travel should be planned for a single person with a budget of \$1,700. &\textbf{CS-Micro}: The proportion of elements in a generated travel plan that do not demonstrate a commonsense error (e.g. visiting the same attraction twice).\newline \textbf{H-Micro}: The proportion of elements in a generated travel plan that do not violate a constraint set by the user or a physical constraint (e.g. budget overruns, non-existent restaurants).\newline \textbf{Final}: The proportion of generated travel plans in which there are no exhibited commonsense errors and all constraints are met (i.e., valid travel plans).\\
\midrule
\textbf{WebArena}\newline \cite{zhou2024webarena}      & test & 271 & Show me the ergonomic chair with the best rating &\textbf{SR}: Whether the task is successfully completed or correctly marked as unachievable.\newline \textbf{SR (AC)}: Whether the task is successfully completed, only among tasks that are achievable.\newline \textbf{SR (UA)}: Whether the task is correctly marked as unachievable, only among tasks that are unachievable.\\
\bottomrule
\end{tabularx}
\caption{The dataset split, number of queries, and example queries from each of the benchmarks we test.}
\label{tab:benchmark_examples}
\end{table}

\section{Additional Design Notes}

\subsection{Prompts}
In this section, we provide the prompts used for each benchmark. We use zero-shot prompts for each benchmark, and provide the necessary tools as defined in each benchmark's paper.

\begin{table}[h]
\centering 
\small
\begin{tabular}{p{0.15\linewidth}|p{0.80\linewidth}} 
\toprule
&\textbf{Prompt}\\
\midrule
\textbf{FanOutQA}\newline \cite{zhu2024fanoutqamultihopmultidocumentquestion}&\texttt{USER: \{question\}}\\
\midrule
\textbf{TravelPlanner}\newline \cite{xie2024travelplanner}&\texttt{SYSTEM: Based on the user's query, make the best travel plan for the user and save it. Do not ask follow-up questions.\newline
USER: \{question\}}\\
\midrule
\textbf{WebArena}\newline \cite{zhou2024webarena}&\texttt{SYSTEM: You are an autonomous intelligent agent tasked with navigating a web browser. You will be given web-based tasks. These tasks will be accomplished through the use of specific functions you can call. \newline
Here's the information you'll have:\newline
The user's objective: This is the task you're trying to complete.\newline
The current web page's accessibility tree: This is a simplified representation of the webpage, providing key information.\newline
The current web page's URL: This is the page you're currently navigating.\newline
The open tabs: These are the tabs you have open.\newline}
\texttt{Homepage:
If you want to visit other websites, check out the homepage at http://homepage.com. It has a list of websites you can visit.\newline}
\texttt{USER: BROWSER STATE: \{observation\}\newline
URL: \{url\}\newline
OBJECTIVE: \{objective\}}\\
\bottomrule
\end{tabular}
\caption{The prompts used for each benchmark in our evaluation.}
\label{tab:benchmark_metrics}
\end{table}

\subsection{Identical Delegation Prevention}
By default, the delegation schemes bundled in ReDel will prevent an agent from delegating instructions that are the same as the instructions that were given to it. If an agent attempts to do so, the delegation function returns a message instructing the agent to either attempt the task itself or break it into smaller pieces before delegating again. We implemented this as an early mitigation for undercommitment, but some undercommitment still occurs.

\end{document}